\documentclass{article}

\usepackage{array}
\usepackage{arxiv}
\usepackage{vcell}
\usepackage[utf8]{inputenc} 
\usepackage[T1]{fontenc}    
\usepackage[colorlinks=true, 
            linkcolor=blue, 
            citecolor=blue, 
            urlcolor=red]{hyperref}
\usepackage{url}            
\usepackage{booktabs}       
\usepackage{amsfonts}       
\usepackage{nicefrac}       
\usepackage{microtype}      
\usepackage{graphicx}
\usepackage{lipsum}
\usepackage{natbib}
\usepackage{amsmath}
\usepackage{amssymb}
\usepackage{amsthm}
\usepackage{times}
\usepackage{wrapfig}
\usepackage{float}
\usepackage{multirow}
\usepackage{array}
\usepackage{makecell}
\usepackage{algcompatible}
\usepackage{algorithm}
\usepackage{caption}
\usepackage{subcaption}

\usepackage{latexsym}
\usepackage{enumitem}
\usepackage{color}
\usepackage{algpseudocode}
\usepackage[justification=justified,singlelinecheck=false]{caption}
\usepackage{ragged2e}

\theoremstyle{definition}
\newtheorem{definition}{Definition}[section]

\title{Mechanism Learning: reverse causal inference in the presence of multiple unknown confounding through causally weighted Gaussian mixture models}

\author{
  Jianqiao Mao \\
  School of Computer Sciencece\\
  University of Birmingham\\
  Birmingham, B15 2TT \\
  \texttt{jxm1417@student.bham.ac.uk} \\
   \And
  Max A. Little \\
  School of Computer Sciencece\\
  University of Birmingham\\
  Birmingham, B15 2TT \\
  \texttt{maxl@mit.edu} \\
}

\renewcommand{\headeright}{}
\renewcommand{\undertitle}{}
\renewcommand{\shorttitle}{\textit{Mechanism Learning}}

\begin{document}
\maketitle
\begin{abstract}%
A major limitation of machine learning (ML) prediction models is that they recover associational, rather than causal, predictive relationships between variables. In high-stakes automation applications of ML this is problematic, as the model often learns spurious, non-causal associations. This paper proposes \emph{mechanism learning}, a simple method which uses \emph{causally weighted Gaussian Mixture Models} (CW-GMMs) to deconfound observational data such that any appropriate ML model is forced to learn predictive relationships between effects and their causes (reverse causal inference), despite the potential presence of multiple unknown and unmeasured confounding. Effect variables can be very high-dimensional, and the predictive relationship nonlinear, as is common in ML applications. This novel method is widely applicable, the only requirement is the existence of a set of mechanism variables mediating the cause (prediction target) and effect (feature data), which is independent of the (unmeasured) confounding variables. We test our method on fully synthetic, semi-synthetic and real-world datasets, demonstrating that it can discover reliable, unbiased, causal ML predictors where by contrast, the same ML predictor trained naively using classical supervised learning on the original observational data, is heavily biased by spurious associations. We provide code to implement the results in the paper, online.
\end{abstract}


\section{Introduction}

        \footnotetext{The experiment code, mechanism learning Python package installation and usage instructions for this study can be accessed through the \href{https://github.com/JianqiaoMao/mechanism-learn}{GitHub repository}.}

    \begin{figure}[htbp]
        \centering
        \includegraphics[width=0.98\textwidth]{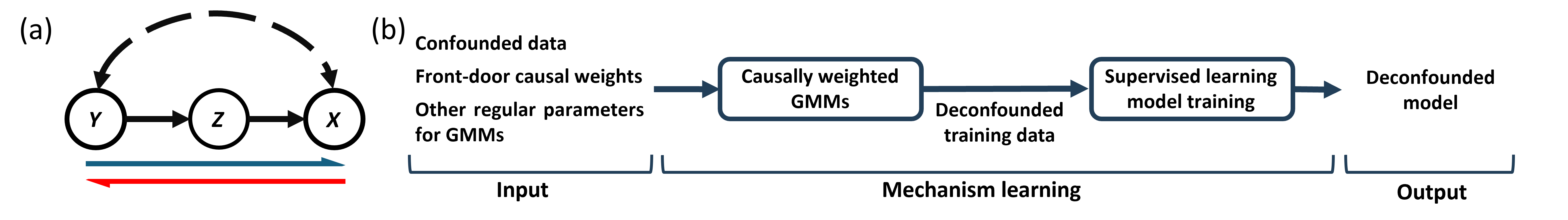}
        \caption{Mechanism learning (b) is a novel, simple and widely applicable solution to the problem of reverse causal inference in the presence of multiple unknown confounding, using arbitrary supervised ML algorithms to predict nonlinear effect-cause relationships from potentially high-dimensional effects. The causal scenario is represented by the ubiquitous \textit{front-door} causal graph (a). There are multiple, unmeasured/unknown confounding paths between $Y$ and $X$ (bi-directed, dashed arrow). The classic causal inference direction is the causal path from $Y$ to $X$ via $Z$ (blue half arrow); reverse causal inference infers causes $Y$ from effects $X$ (red half arrow).}
        \label{fig: mechanism learning}
    \end{figure}

    In recent years, machine learning (ML) has shown promising performance in many applications, to name but a few: disease diagnosis \citep{ahsan2022machine}, automatic driving \citep{alam2021review}, and adaptive control \citep{mao2023data, zamfirache2023neural}. However, one of the major limitations of applying ML methods in critical, high-stakes applications such as decision-making in medicine, is their ``causal blindness'', that is, ML models are, by design, pattern-recognition algorithms which learn potentially spurious, non-causal associations between the feature and target variables. 

    One solution to this blindness is to train ML algorithms on experimental (interventional) data which eliminates unwanted associations, capturing only the desired causal relationships. Unfortunately, such data, collected under well-controlled experimental settings, is normally expensive and comparably small-scale, which may not meet the heavy sample size demands of modern, sophisticated ML models \citep{zhao2023survey}. Moreover, controlled experiments may be physically impossible or unethical to conduct, for instance in global climate science and macroeconomics. By contrast, observational data is now easier and cheaper to collect due to advances in data collection, storage and sensing technology. Nevertheless, observational data can be easily corrupted by known or latent confounding factors \citep{ming2022impact}. Interestingly, the prediction performance of ML models trained on confounded data may be \emph{significantly over-estimated} and fail to hold when confounders change or are no longer present, as often happens in practical deployment settings. The primary reason for this overestimate is simple -- association does not generally imply causation -- except under special circumstances. In terms of Pearl's \emph{do-calculus}, $p\left(\left.x\right|do\left(y\right)\right) \neq p\left(\left.x\right|y\right)$ for most situations where observational data alone is available \citep{pearl2009causality}.

    Various \emph{causal inference} (CI) methods have been proposed to identify and quantify causal effects from observational data, such as instrumental variables \citep{baiocchi2014instrumental}, propensity score matching \citep{yang2021propensity} and advanced ML algorithms such as the causal forest predictor \citep{wager2018estimation}. A systematic approach to nonparametric structural causal inference involves the integration of \emph{structural causal models} (SCM), \emph{do-calculus} and related analytical methods \citep{pearl2009causality, richardson2023nested}. 
    
    These existing CI methods, although useful, are generally not applicable to high-dimensional variables or estimating nonlinear relationships, while ML predictors are. Supervised ML predictions are typically made in the \emph{reverse causal} direction, that is, counter to the generative causal arrow, predicting causes from effects, unlike traditional CI where effects of causes are predicted (for instance, average treatment effects). For example, in the front-door setting (Figure \ref{fig: mechanism learning}(a)), the cause $Y$ can be some diagnostic category and the effect $X$ may be some high-dimensional features such as images or other digital measurements. The ML prediction goal is to use the features to predict the diagnostic category. Here, the mechanism variable $Z$ could represent some \emph{disease-dependent mechanism} that ``explains'' the diagnostic category for given digital images. Meanwhile, a set of unobserved confounders may exist (e.g., patient demographics), which introduces spurious correlations between $X$ and $Y$.
    
    Ideally, we want to sample from the interventional distribution $p\left(\left.x\right|do\left(y\right)\right)$ to obtain a non-confounded dataset for causal ML predictor training, so that the model learns the causal relationship and thus makes causal predictions unbiased to confounding effect. However, the interventional distribution, in general, does not always have a closed form, which makes directly sampling from the interventional distribution of interest difficult and even impossible. \cite{little2019causal} presented the \emph{causal bootstrapping} (CB), a weighted bootstrap which deconfounds the observational data so as to approximate the distribution of non-confounded, interventional data. The resampled (deconfounded) data can be used to train a causal ML predictor which learns the intended, unbiased and causal feature-target relationships. However, the simple bootstrapping method tends to generate many replicas due to the extreme CB weight distribution. And thus, a reduction of the sample variability can be witnessed, which potentially leads to underfitting problems when it is applied to high-dimensional and complicated tasks.

    This report proposes \emph{mechanism learning}, a novel method that enables arbitrary supervised ML models to perform reverse causal inference for the front-door confounding scenario without altering the ML model. Such an ML model trained using mechanism learning, will be an estimate of the true, causal, feature-target relationship, \emph{unbiased by any spurious associations} between these variables. Mechanism learning is widely applicable: specifically, to any situation where, in addition to label and feature data, there exists data for the mechanism. For instance, in radiology, disease-specific morphological features may mediate the effect of diagnostic categories on images, while multiple unmeasured confounders (e.g., imaging protocols, patient demographics) influence both. As illustrated in Figure~\ref{fig: mechanism learning}, mechanism learning resamples observational data using \emph{causally weighted Gaussian Mixture Models} (CW-GMMs), where the causal weights are derived from the front-door interventional distribution inform the sampling. Compared to nonparametric CB resampling, CW-GMMs provide better sample variability and can generate \emph{i.i.d.} deconfounded samples from the approximated interventional distribution.

    To demonstrate effectiveness, we first train simple ML models using mechanism learning on synthetic and semi-synthetic datasets with known generative processes, clearly showing superior robustness to confounding compared to classical supervised learning. Next, we apply mechanism learning to a complex real-world dataset—intracranial hemorrhage (ICH) detection from computed tomography (CT) images \citep{hssayeni2020computed}. Results show that classical supervised learning severely overfits due to unknown confounders, causing performance degradation when these confounders are removed by CW-GMMs. In contrast, ML models using mechanism learning show stable performance across confounded and non-confounded settings and consistently outperforms the existing CB-based deconfounding method. This strongly suggests that mechanism learning enables the ML model to capture the causal feature-target relationship rather than spurious associations.

    Next, we describe the proposed mechanism learning method in detail.

\section{Mechanism learning}

    The core of this method is the use of CW-GMMs applied to the front-door SCM \citep{bellemare2024paper}. Using the front-door weights presented by \cite{little2019causal} (see also Appendix \ref{sec: Appendix B} for theoretical derivations), we can fit CW-GMMs which deconfound the observational data before it is used to train a standard supervised learning model. The following sections present the process of encoding the causal knowledge into CW-GMMs, and show the training procedure of a causal supervised learning predictor using the \emph{mechanism learning}.

    \subsection{Front-door criterion and causal weights}

    A formal definition of the ubiquitous \textit{front-door criterion} is given by the following:

    \begin{definition}[Front-door criterion \citep{pearl2009causality}]
    Given variable sets $Y$, $X$ and $Z$, the causal effect from the cause $Y$ to the effect $X$ is identifiable if:
        \begin{enumerate}
            \item The mechanism $Z$ intercepts all directed paths from $Y$ to $Z$.
            \item There is no unblocked backdoor path from $Y$ to $Z$; and
            \item All backdoor paths from $Z$ to $X$ are blocked by $Y$.
        \end{enumerate}
    \end{definition}

    Importantly, if one can identify a set of such \emph{mechanism} variable set $Z$ satisfying front-door criterion, the number or nature of latent confounders becomes irrelevant: the interventional distribution the causal relationship $p\left(x \mid do(y)\right)$ becomes uniquely computable from observational distributions as shown in \eqref{eq: front-door intv prob}. Selecting the mechanism variable $Z$ should be informed by domain knowledge and the specific application context, while keeping the front-door criterion satisfied. For instance, in healthcare $Z$ might be disease-specific biomarkers that mediate diagnosis and imaging features, while in economics $Z$ could be skill or qualification levels mediating the effect of education on income.

    \begin{equation}
        \label{eq: front-door intv prob}
        p\left(\left. x \right|do\left(y\right)\right) = \int_{z} p\left(\left.z\right|y\right) \int_{y'} p\left(\left.x\right|y',z\right)p\left(y'\right) dy'dz,
    \end{equation}
    where the random variable $Y'$ is a copy of variable $Y$ which arises during the application of \emph{do-calculus}.

    In the study \citep{little2019causal}, the front-door weights are introduced as written in \eqref{eq: front-door cb weights}, which encodes the front-door interventional distribution for the given observations (data set) $\mathcal{D} = \{\left(x_n, y_n, z_n\right)\}_{n=1}^N$ under front-door confounding assumption.

    \begin{equation}
    \label{eq: front-door cb weights}
          w_{n}\left(z_n, y_n, y_i\right) =  \frac{p\left(\left.z_{n}\right|y_i\right)}{N p\left(\left.z_{n}\right|y_{n}\right)}.
    \end{equation}  

    Additionally, the computational efficiency and accuracy of the front-door weights are guaranteed even for relatively small datasets $\mathcal{D}$. This is because we do not need to estimate any marginal or joint distributions associated with the high-dimensional features $\boldsymbol{X}$. However, due to the ratio construction of the front-door weights, it naturally tends to get extreme values when the probability $p\left(z \mid y'\right)$ in the denominator approaches 0. This implies that CB presented in the study \citep{little2019causal} is very likely to generate replicas for a small portion of the original data with relatively large causal weights.

    \subsection{Causally weighted GMMs (CW-GMMs)}

        As aforementioned, even though directly sampling from the interventional distribution is demanding, we can instead sample efficiently from an approximated interventional distribution. With the computed CB weights that embed the causal structure, it is then possible to approximate the identified interventional distribution using an appropriate parametric model.
        
        Gaussian mixture model (GMM) as our approximation with three-fold motivations as follows: First, GMMs serve as universal density approximators, capable of modeling complex multimodal distributions given a sufficient number of mixture components. Second, compared with direct weighted bootstrapping that often suffers from limited sample variability when it encounters extreme weights, the probabilistic nature of GMM-based resampling inherently introduces diversity and smoothness in the samples, mitigating issues of collapsed distributions. Third, GMMs are computationally tractable, and their parameters can be efficiently estimated via well-established optimization techniques.
    
        This inspires the \emph{CW-GMM}, which extends the standard GMM by modifying the optimization objective through incorporating sample-specific causal weights. Specifically, we construct the mixture coefficients $\pi_{k}$, component means $\boldsymbol{\mu}_{k}$, covariances $\boldsymbol{\Sigma}_{k}$, and the log-likelihood function $\log \mathcal{L}(\theta)$ in a causally weighted form. Algorithm \ref{alg:weighted-gmm} outlines the core steps of the implemented CW-GMM optimization process using the \emph{EM algorithm} \citep{collins1997algorithm}. For clarity, we omit details in Algorithm \ref{alg:weighted-gmm} such as covariance regularization, log-sum-exp stabilization in likelihood computation, which are used in our actual implementation for numerical stability. We show some sensitivity analysis for the key hyperparameter, the number of components $K$, in the later sections.
    
        \begin{algorithm}[ht]
        \caption{CW-GMM Optimization using EM Algorithm}
        \label{alg:weighted-gmm}
        \begin{algorithmic}[1]
        \item[] \textbf{Input:} Observational sample of the effect variable $X = \{x_n\}_{n=1}^N \subset \mathbb{R}^d$, causal weights $w = \{w_n\left(z_n, y_n, y_i\right)\}_{n=1}^N$ for a given intervention $y_i$, number of components $K$, maximum iterations $T$
        \item[] \textbf{Output:} A CW-GMM $f\left(x \mid \mathrm{do}(Y=y_i)\right) = \sum_{k=1}^K \pi_k \cdot \mathcal{N}\left(x \mid \mu_k, \Sigma_k\right)$
 
        \State Initialize $\pi_k \gets \frac{1}{K}$, $\mu_k$ via weighted $k$-means++, $\Sigma_k$ via global weighted covariance
        
        \For{$t = 1$ to $T$}
            
            \For{$n = 1$ to $N$} \Comment{E-step}
                \For{$k = 1$ to $K$}
                    \State Compute $\gamma_{nk} \gets \frac{\pi_k  \mathcal{N}\left(x_n \mid \mu_k, \Sigma_k\right)}{\sum_{j=1}^K \pi_j  \mathcal{N}\left(x_n \mid \mu_j, \Sigma_j\right)}$
                \EndFor
            \EndFor
            
            \For{$k = 1$ to $K$} \Comment{M-step}
                \State $W_k \gets \sum_{n=1}^N w_n  \gamma_{nk}$
                \State $\pi_k \gets \frac{W_k}{\sum_{j=1}^K W_j}$
                \State $\mu_k \gets \frac{1}{W_k} \sum_{n=1}^N w_n  \gamma_{nk}  x_n$
                \State $\Sigma_k \gets \frac{1}{W_k} \sum_{n=1}^N w_n  \gamma_{nk}  \left(x_n - \mu_k\right)\left(x_n - \mu_k\right)^\top$
            \EndFor
        
            \If{$|\log \mathcal{L}^{\left(t\right)} - \log \mathcal{L}^{\left(t-1\right)}| < \varepsilon$}
                \State \textbf{break}
            \EndIf
        \EndFor
        \end{algorithmic}
        \end{algorithm}
    
    \subsection{ML predictor training with mechanism learning}

    To train a causally unbiased (deconfounded) ML predictor using \emph{mechanism learning}, we present the model training process as outlined in Algorithm \ref{alg:cb-gmm-deconf}. Given a set of fitted CW-GMM $F = \{f^{\left(m\right)} : m=1,2,...,M\}$, where $M$ is the number of intervention values and $f^{\left(m\right)}$ is the $m$-th CW-GMM fitted with the front-door weights $w^{\left(m\right)} = \left\{w_n^{\left(m\right)} \left(z_n, y_n, y^{\left(m\right)} \right) \right\}_{n=1}^N$ with respect to the interventional value $y_i = y^{\left(m \right)} \in \Omega_{I}$, such that $\Omega_{I}$ is the sample space of the intended intervention. This process allows the ML predictor to learn the true causal relationship between $X$ and $Y$ without ever observing actual interventional data — only using the deconfounded samples that approximate the interventional distribution. So far, the trained model using the presented strategy can make causally unbiased predictions, which is referred to as \emph{mechanism learning} in the previous discussion.
    
    \begin{algorithm}[htbp]
        \caption{Training the ML Predictor for Mechanism Learning via CW-GMM Resampling}
        \label{alg:cb-gmm-deconf}
        \begin{algorithmic}[1]
        \item[] \textbf{Input:} A set of CW-GMM $F = \{f^{\left(m\right)} : m=1,2,...,M\}$ corresponding to each interventional value $y^{\left(m \right)} \in \Omega_{I}$, sample sizes $\{r^{\left(m\right)}\}$
        \item[] \textbf{Output:} A deconfounded ML model $h: \mathcal{X} \rightarrow \mathcal{Y}$
        
        \For{each intervention value $y^{\left(m\right)}$}
            \State Sample from $f^{(m)}$ to obtain $\tilde{X}^{(m)} \gets \left\{ \tilde{x}_i^{(m)} \right\}_{i=1}^{r^{\left( m \right)}}$
        \EndFor
        
        \State Form the deconfounded dataset:
        \Statex \hfill $\tilde{\mathcal{D}} \gets \bigcup_{m=1}^M \left\{(\tilde{x}_i^{(m)}, y^{(m)})\right\}_{i=1}^{r^{\left(m\right)}}$ \hfill~
        \State Train model $h$ on the deconfounded dataset $\tilde{\mathcal{D}}$.
        \end{algorithmic}
    \end{algorithm}

\section{Tests with fully- and semi-synthetic generative models}

    Throughout experiments, we define \emph{confounded data} as observational data influenced by unmeasured confounders; \emph{non-confounded data} as data collected from controlled or randomized experiments; and \emph{deconfounded data} as data generated from observational data using causal resampling techniques (e.g., CW-GMM or CB), approximating the interventional distribution. As for the ML models trained with these data, we call them \emph{confounded models} for those trained on confounded data, \emph{deconfounded models} for those trained on deconfounded data. Specifically, a supervised ML model trained on the CW-GMM-based deconfounding method is called the \emph{mechanism learning-based deconfounded model}, while a supervised ML model trained on the CB-based deconfounding method is called the \emph{CB-based deconfounded model}.

    For all simulations in this section, we compare three models: (i) mechanism learning-based deconfounded model (as Algorithm \ref{alg:cb-gmm-deconf}), (ii) CB-based deconfounded model, and (iii) a confounded model. All performance evaluations are based on both non-confounded and confounded test data.

    Two fully synthetic datasets were generated for classification (categorical prediction target $Y$) and regression (continuous prediction target $Y$). For the semi-synthetic dataset, we modify the MNIST dataset \citep{lecun1998gradient} by adding a background brightness confounder under the front-door causal setting, and a proxy binary mechanism variable that mediates the causal effect from digit categories $Y$ to the handwriting images $X$. For generative model details, see Appendix \ref{sec: Appendix A}.

\begin{figure}[htbp]
        \centering
        \includegraphics[width=0.9\columnwidth]{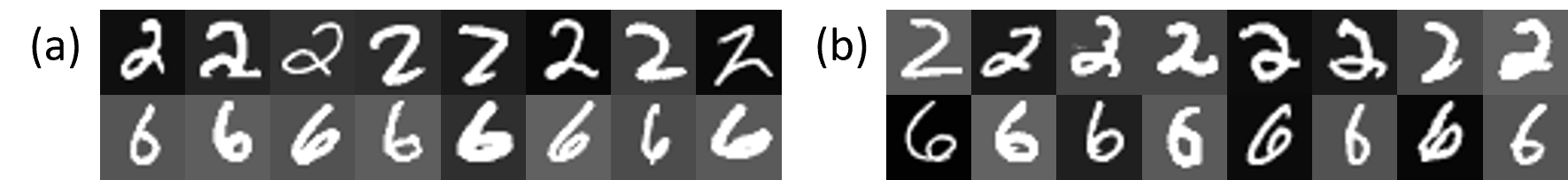} 
        \caption{Example digits from the confounded (a) and non-confounded (b) background-MNIST datasets. In (a), background brightness is manipulated so that it is a confounding factor with digit class (e.g., "6" is brighter than "2"); in (b), the brightness-digit association is randomized to simulate a controlled setting.}
        \label{fig: brightness MNIST}
    \end{figure}

    As a showcase, we discuss the semi-synthetic background-MNIST dataset to demonstrate how unmitigated confounding in the training data can lead to a causally biased and therefore unreliable, ML model. Note that, this bias cannot be detected from investigation of the confounded test set accuracy alone. Instead, a causal-aware ML model should have certain robustness in terms of model performance when confounders are no longer present or changed. In the confounded dataset, images of digit ``6'' tend to have brighter backgrounds than images of digit ``2'', but this confounding effect is not present for the non-confounded dataset (Figure \ref{fig: brightness MNIST}). A standard supervised classifier trained on the confounded MNIST dataset is likely to make predictions based on the input image's average brightness rather than the handwriting digit's actual shape because the brightness feature is strongly associated with the label. Thus, any supervised learning algorithm will use this spurious brightness information to maximize prediction accuracy. However, this brightness information is not what we expect the classifier to learn; should the predictor be applied to data without this brightness confounder, it would be of no use, despite the apparently high out-of-sample accuracy of the predictor.

    \subsection{Test results: synthetic classification dataset}

\begin{figure}[ht]
        \centering 
        \includegraphics[width=0.98\columnwidth]{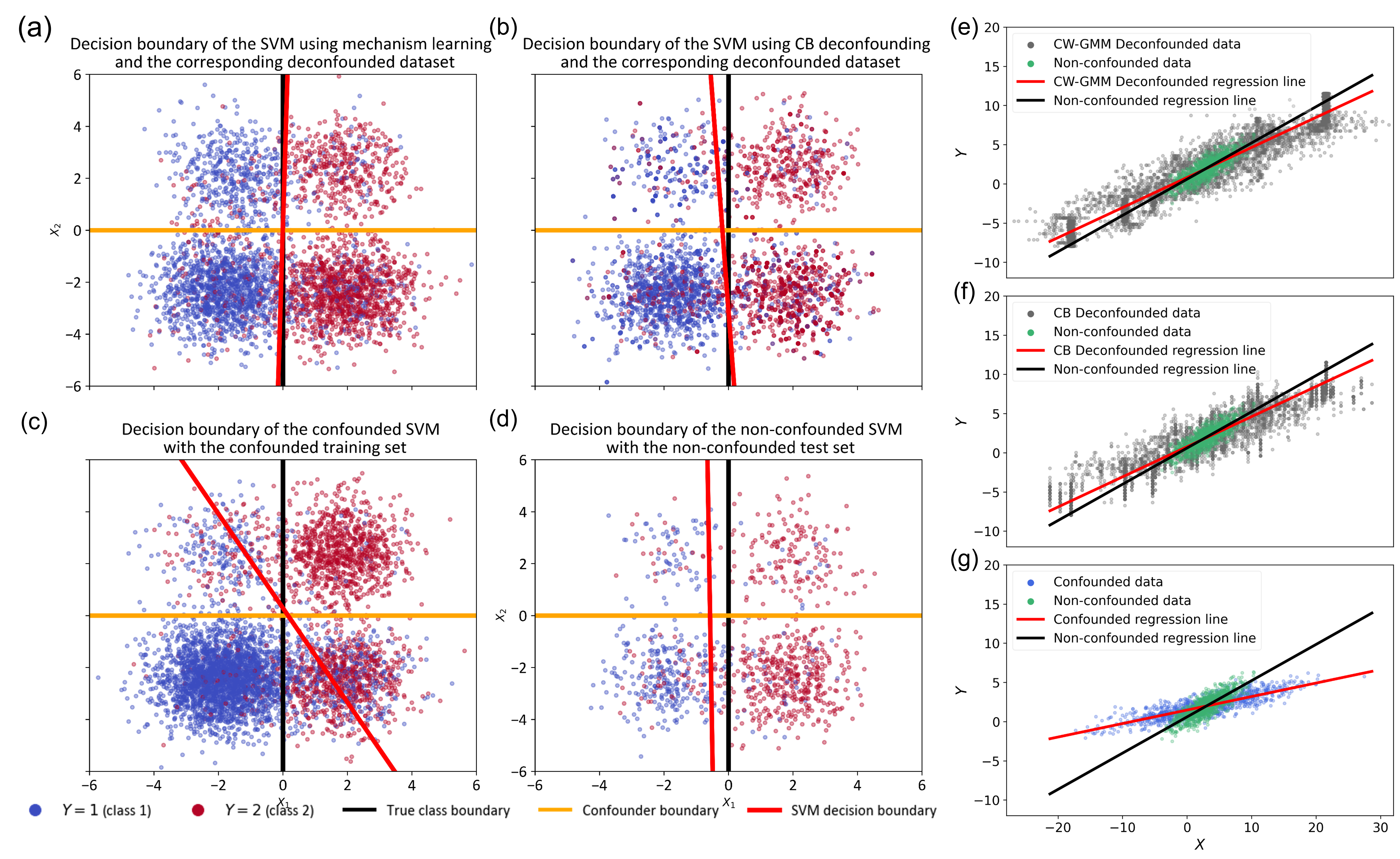}
        \caption{Comparison of classification (a-d) and regression (e-g) models trained using mechanism learning (a, e), CB-based deconfounding (b, f), and classical supervised learning trained on confounded data set (c, g) and on non-confounded data set (d). 
        In classification (a–d), the confounded model (c) yields a skewed decision boundary, which is a mixture of the confounder (orange line) the true class boundary (black line). Nevertheless, mechanism learning-based deconfounded SVM (a) and CB-based deconfounded SVM (b) produce boundaries aligned with the true class separation, which are closer to the non-confounded SVM boundary shown in (d). The samples generated by CW-GMMs show better sample diversity.
        In regression (e–g), the classical model (g) shows a biased slope due to latent confounding, while mechanism learning (e) and CB-based deconfounding (f) recover the non-confounded regression line (black lines).  }
        \label{fig: syn test results}
    \end{figure}

    Using a simple linear-kernel support vector machine (SVM) classifier, we compare predictive decision boundaries of the mechanism learning-based deconfounded (Fig. \ref{fig: syn test results} (d)), CB-based deconfounded (Fig. \ref{fig: syn test results} (e)), and confounded SVMs (Fig. \ref{fig: syn test results} (f)). The confounded SVM, trained using classical supervised learning, is severely affected by the confounder, whose decision boundary conflates the true class with the implied confounder boundaries. By contrast, the decision boundaries of the mechanism learning-based and CB-based deconfounded SVMs closely match the true class boundary, which evidences that they nullify the influence of the confounder. Therefore, both of the deconfounded models capture the desired causal relationship between features and prediction target. The blue and red dots represent the originally confounded dataset in Fig. \ref{fig: syn test results} (f) and the deconfounded datasets generated by the CW-GMM sampler used for mechanism learning in Fig. \ref{fig: syn test results} (d) and by the CB sampler used for CB-based deconfounded model training in Fig. \ref{fig: syn test results} (e). Noticeably, the CW-GMM sampler generates new samples with better variability compared to the CB sampler, though both of the deconfounded models show satisfying deconfounding performance because of the simplicity of the synthetic dataset.

    In terms of performance evaluation results (Table \ref{tab: expr results}), the mechanism learning-based deconfounded model and CB-based deconfounded model exceed the confounded SVM on the non-confounded dataset by $7\%$ accuracy, retaining stable predictive accuracy across both confounded and non-confounded datasets. On the contrary, although the confounded SVM performs about as well as the mechanism learning-based SVM in the confounded dataset, its performance declines significantly on the non-confounded dataset. This is because classical supervised learning is biased by the influence of confounding, reporting misleading accuracy on the original (confounded) test set. However, due to the simplicity and low dimensionality of the synthetic data, the performance of CB-based and mechanism learning-based deconfounded models remains largely indistinguishable.

    The CW-GMM is inherently moderately sensitive to the number of Gaussian components, and the selection of this key hyperparameter $K$ largely depends on the data distributions in a specific context. Nevertheless, there are a number of standard hyperparameter tuning methods for empirical analysis, such as the Akaike information criterion (AIC) and the Bayesian information criterion (BIC) for pre-screening \citep{chakrabarti2011aic}.

\begin{figure}[htbp]
        \centering 
        \includegraphics[width=0.98\columnwidth]{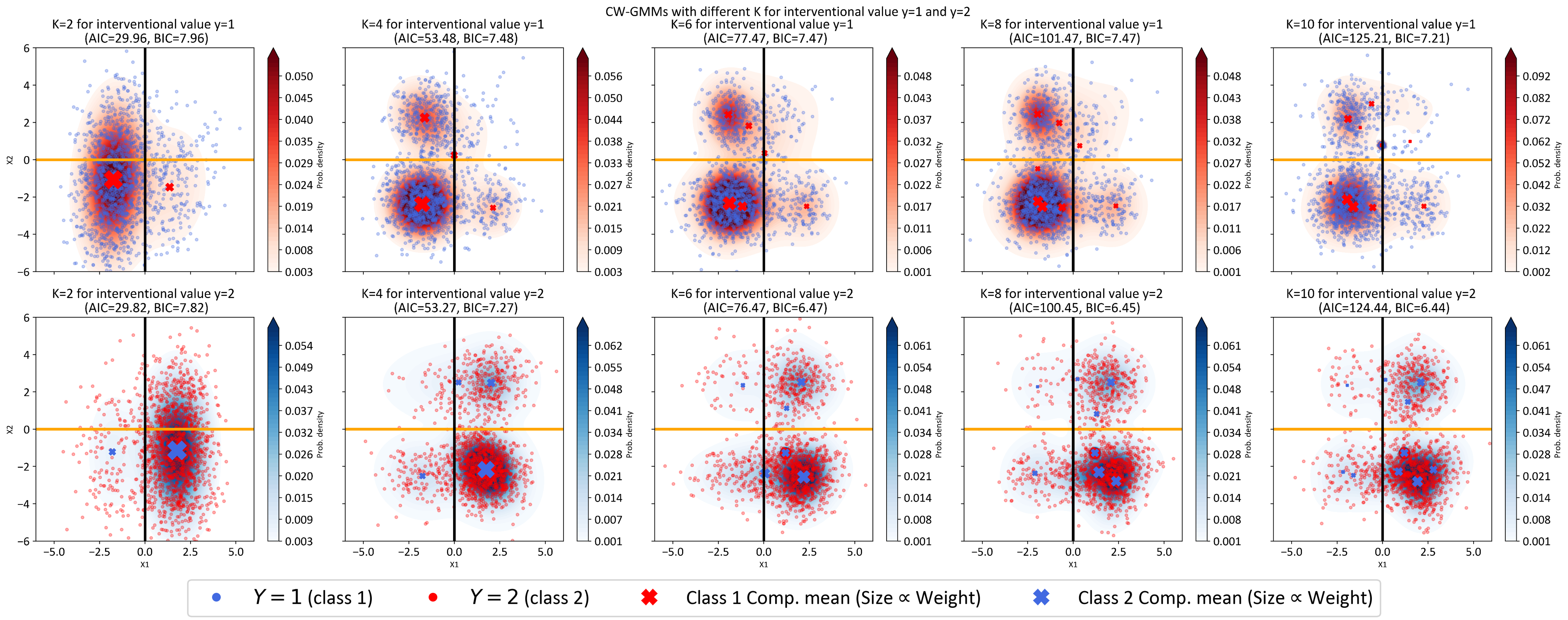}
        \caption{Comparison of the fitted CW-GMMs with different numbers of components $K$ (2, 4, 6, 8 and 10 Gaussian components, respectively) for the synthetic classification task, together with the deconfounded data generated correspondingly. For each intervention, a CW-GMM is fitted. The mark size of the component mean is proportional to the mixture coefficient $\pi_k$, and the higher probability density area is reflected by a darker color.}
        \label{fig: diff K CWGMM}
    \end{figure}

    For demonstration, Figure \ref{fig: diff K CWGMM} compares the total probability density together with the component means of the fitted CW-GMMs with different hyperparameters $K$. Although the choice of $K$ substantially determines the complexity of the fitted CW-GMMs, the generated samples closely approximate the intended interventional distribution. We attribute this robustness to the causal weights: because the causal weights inform the key statistics, such as mixture coefficient, component means, etc., a few components incorporating the causal structure dominate the model, which gives a reasonable safety margin in choosing $K$ in practice.

    \subsection{Test results: synthetic regression dataset}

    Training three simple univariate linear regression models, we again compare the predictive performance of confounded (Figure \ref{fig: syn test results} (a)), CB-based deconfounded (Figure \ref{fig: syn test results} (b)) and mechanism learning-based deconfounded models (Figure \ref{fig: syn test results} (c)). In this regression problem, both deconfounded datasets are sampled using a uniformly distributed interventional variable, rather than by matching the distribution of the observational cause variable. Meanwhile, the interventional values are from an extended sample space $\hat \Omega_{Y}$ compared to the observational cause value space $\Omega_{Y}$, which is considered the so-called counterfactual intervention. Classical regression leads to a model whose regression line has a much smaller slope than the true, causal slope between feature $X$ and the target $Y$ (See Appendix \ref{subsec: Appendix A.2} for the ground truth of the confounded and non-confounded relationship in the data generation process).

    In terms of prediction error (Table \ref{tab: expr results}), the mechanism learning-based deconfounded model has a smaller error (by around $0.4$ units in mean square error) than the confounded regression model, when both are applied to the non-confounded test dataset. Although the confounded linear regression model exceeds the performance of the mechanism learning-based deconfounded one on the confounded test set, this is not a stable and reliable regression relationship which significantly depends on the confounding factor. Meanwhile, this is not the desired true causal relationship that we wish the model to learn. Similar to the simulations run on the synthetic classification dataset, the performance difference remains subtle between the mechanism learning-based and CB-based deconfounded models, which is, again, because of the simplicity of the dataset itself. We expect to observe more differentiable performance metrics in the following experiments on datasets with high-dimensional, complicated, non-linear feature-target relationships.

    \subsection{Test results: semi-synthetic background-MNIST dataset}

    A simple ML classification model (K-nearest neighbors, KNN) achieves fairly high prediction accuracy at separating digits (Table \ref{tab: expr results}) on background-MNIST data, regardless of whether the ML model is trained using classical supervised learning, mechanism learning or CB-based deconfounding. However, on non-confounded test data, the mechanism learning-based deconfounded model exceeds the performance of the confounded model by a significant margin and slightly outperforms the CB-based deconfounded model. This is good evidence that the classical model is heavily biased by the strong confounding, i.e., the background brightness. By contrast, the mechanism learning-based deconfounded model makes unbiased causal predictions about the shape of the digits, which are robust to background brightness. Compared to the CB-based deconfounding method, the presented mechanism learning trains a ML predictor making more accurate causal predictions on the non-confounded test dataset. We deduce this because CW-GMM resampling enhances the sample variability, whereas CB reduces the variability.

        \begin{table*}[htbp]
        \begin{centering}
        \small%
        \begin{tabular*}{1\textwidth}{@{\extracolsep{\fill}}>{\raggedright}p{0.19\textwidth}>{\raggedright}m{0.13\textwidth}>{\raggedright}p{0.15\textwidth}>{\raggedright}m{0.1\textwidth}>{\centering}p{0.15\textwidth}>{\centering}p{0.15\textwidth}}
        \toprule 
        \multirow{2}{0.19\textwidth}{Task} & \multirow{2}{0.13\textwidth}{ML Model} & \multirow{2}{0.15\textwidth}{Deconfounding method} & \multirow{2}{0.1\textwidth}{Performance metric} & \multicolumn{2}{c}{Test data}\tabularnewline
        \cmidrule{5-6} \cmidrule{6-6} 
         &  &  &  & Non-confounded & Confounded\tabularnewline
        \midrule
        \midrule 
        \multirow{3}{0.19\textwidth}{Synthetic classification} & \multirow{3}{0.13\textwidth}{Linear SVM} & / & \multirow{3}{0.1\textwidth}{Accuracy (\%)} & 78.20 & 85.30\tabularnewline
         &  & CB &  & 85.40 & 85.50\tabularnewline
         &  & CW-GMM (Ours) &  & 85.40 & 85.60\tabularnewline
        \midrule
         & \multirow{3}{0.13\textwidth}{KNN} & / & \multirow{3}{0.1\textwidth}{Accuracy (\%)} & 75.64 & 97.03\tabularnewline
        Background-MNIST  &  & CB &  & 93.11 & 95.23\tabularnewline
        (semi-synthetic) &  & CW-GMM (Ours) &  & 94.07 & 95.13\tabularnewline
        \midrule
         & \multirow{3}{0.13\textwidth}{ResNet-CNN} & / & \multirow{3}{0.1\textwidth}{F1 score} & 0.8401 & 0.9353\tabularnewline
        ICH detection &  & CB &  & 0.9003 & 0.8823\tabularnewline
        (Real-world application) &  & CW-GMM (Ours) &  & 0.9760 & 0.9176\tabularnewline
        \midrule
        \multirow{3}{0.19\textwidth}{Synthetic regression} & \multirow{3}{0.13\textwidth}{Linear Regression} & / & \multirow{3}{0.1\textwidth}{MSE} & 1.162 & 0.595\tabularnewline
         &  & CB &  & 0.662 & 2.679\tabularnewline
         &  & CW-GMM (Ours) &  & 0.658 & 2.649\tabularnewline
        \bottomrule
        \end{tabular*}\normalsize
        \par\end{centering}
        \caption{Test results comparing classical supervised learning and CB-based deconfounding against mechanism learning on fully-, semi-synthetic and real-world ICH detection datasets, where the fully- andd semi-synthetic data is generated from known structural models while the generative structure is unknown for the real-world ICH detection data.
        \label{tab: expr results}}
        \vspace{-1em}
        \end{table*}

\section{Application to real-world data: ICH detection with CT scans}
\label{Sec: real-world}

    The above simulation results show that \emph{mechanism learning} overcomes the failure of confounding bias when classical supervised learning models are trained on confounded, observational data to a significant degree. We now turn to tests on real-world data, where the data-generating process is unknown.

    ICH is a life-threatening medical condition but difficult and time-consuming to diagnose \citep{caceres2012intracranial}. The current clinical approach to ICH detection is based on radiologist examination of CT scans. Many efforts have been made to develop automatic, efficient, ML-assisted tools for the ICH diagnosis \citep{hssayeni2020computed, wang2021deep}. However, data from CT scans labeled with an ICH diagnostic category collected as a result of routine clinical practice, can be severely confounded by many unmeasurable or unknown factors, such as the imaging device's hardware or settings, hospital usage protocols, the patient's age or sex, and combinations of all the above factors. Developing reliable ML-based detection algorithms using this dataset is difficult as a result.

   \begin{figure}[htbp]
        \centering
        \includegraphics[width=0.75\columnwidth]{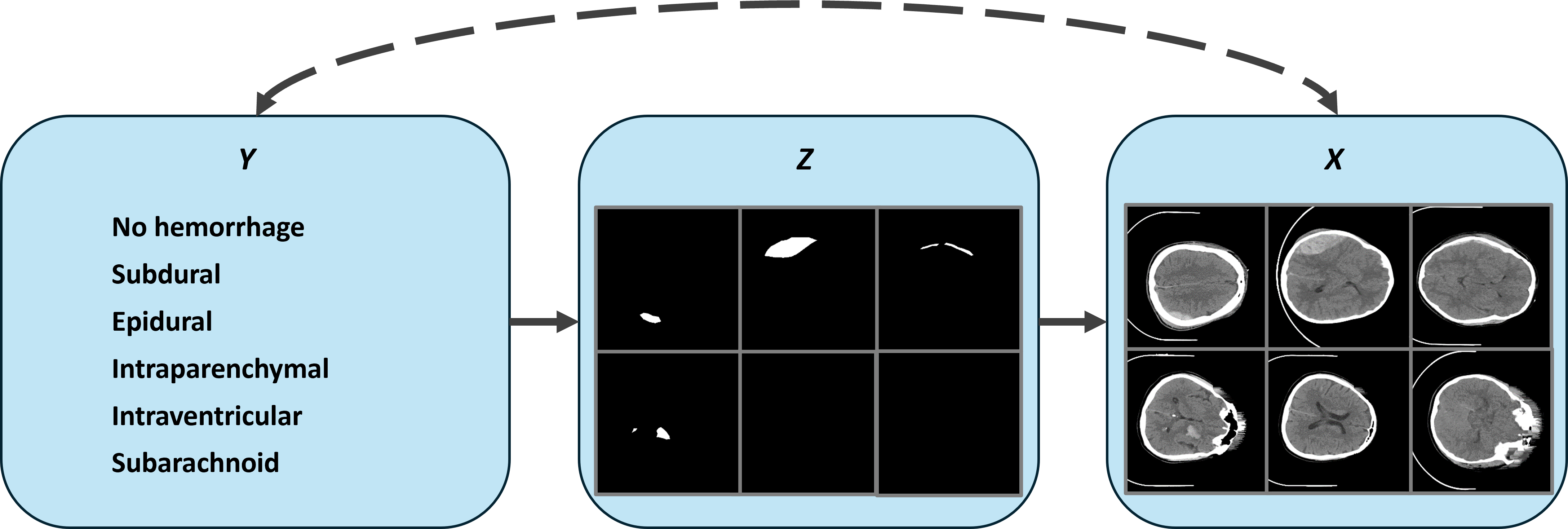}
        \caption{Front-door structural causal model for the real-world ICH dataset \citep{hssayeni2020computed}, for the purposes of mechanism learning. The cause variable $Y$ represents diagnostic category; mechanism variable $Z$ represents hemorrhage region label, and the effect variable $X$ are the digital CT scans.}
        \label{fig: mechanism learning for ICH}
    \end{figure}  

    \subsection{Dataset description and processing}

    For this dataset, the cause variable is the six-class diagnostic ICH category, the mechanism variable is the segmentation of the potential hemorrhage region in CT scans, and the effect variable is high-dimensional CT images (Figure \ref{fig: mechanism learning for ICH}).
    
    There are 82 sets of CT scans including 36 slices for each set. There are five abnormal ICH categories: intraventricular, intraparenchymal, subarachnoid, epidural and subdural, plus a normal category representing no hemorrhage. Each slice is consensus region-labeled by two radiologists, annotating the hemorrhage type. The radiologists have no access to the clinical history of the patients. 
    
    All the original CT scan slices and labeled hemorrhage region images are $512 \times 512$ pixel grayscale images whose pixel value range is $\left[0, 255 \right]$. These are downsampled to $128 \times 128$ pixels to save computational effort. Furthermore, we clean CT scan slices by a semi-manual selection process: if a slice's black area is more than $90\%$, they are filtered out; if a slice's black area is between $80\% - 90\%$, we determine if this slice should be kept by screening on a one-by-one basis.

    A severe imbalance of diagnosis category is observed, where most of the slices are labeled as ``no hemorrhage'' (more than $80\%$) and some abnormal categories have very few samples (less than $2\%$). Therefore, we rebalance the training sample sizes based on class in the following way. For the deconfounded data, we generate 5000 samples for each class (intervention) from the CW-GMMs and CB. To ensure fair comparisons, random oversampling is used to resample the confounded dataset, keeping exactly the same size for each category as the deconfounded training data.

    \subsection{Mechanism embedding}

    \begin{figure}[htbp]
        \centering
        \includegraphics[width=0.65\columnwidth]{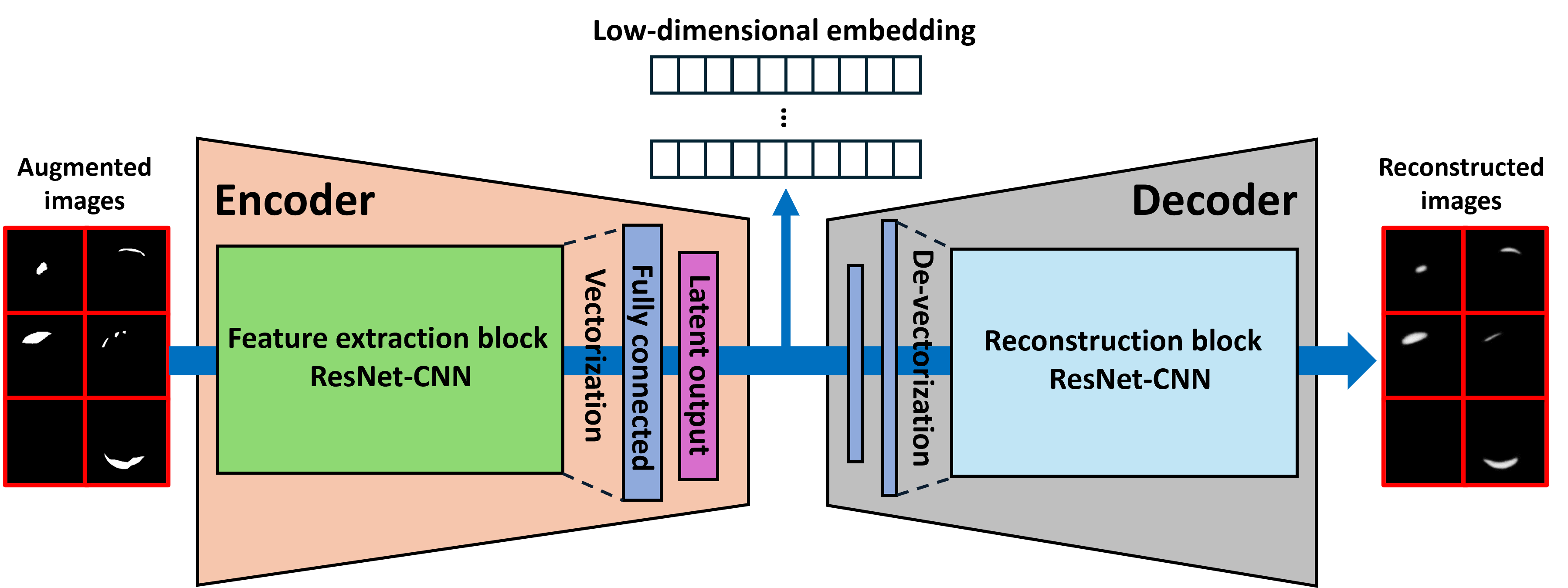}
        \caption{Encoder-decoder architecture for mechanism variable embedding. A 5-layer ResNet-CNN encoder maps hemorrhage region images to a 10-dimensional latent space. The decoder mirrors the encoder, replacing pooling layers with convolutional layers image reconstruction.}
        \label{fig: autoencoder structure}
        \vspace{-0.5em}
    \end{figure} 

    To apply mechanism learning, we need to estimate the joint distribution of the mechanism and cause variables, $p\left(z,y\right)$. However, accurate distribution estimation for such a high-dimensional mechanism variable is impractical. Therefore, we design a ResNet-based encoder-decoder model to extract a 10-dimensional embedding from the labeled ICH regions (Figure \ref{fig: autoencoder structure}). A combined loss function takes the weighted sum of the pixel-wise cross-entropy loss and structure similarity loss. Data augmentation is applied to input images by random rotation, shifting, flipping and zooming.

    \subsection{Non-confounded test data synthesis and evaluation}

    Unlike the case with synthetic data, we do not have the non-confounded data so as to make model performance comparisons. Instead, we assume that the CW-GMM sampler produces deconfounded data that mimics real-world ICH data without confounders. Samples drawn from the CW-GMMs are \emph{i.i.d}, and thus it is easy to form a deconfounded training dataset and a synthesized ``non-confounded'' test dataset that are mutually disjoint. To retain consistency, the synthetic ``non-confounded'' test set contains the same number of samples as the confounded test set.

\begin{figure}[htbp]
        \centering 
        \includegraphics[width=0.65\columnwidth]{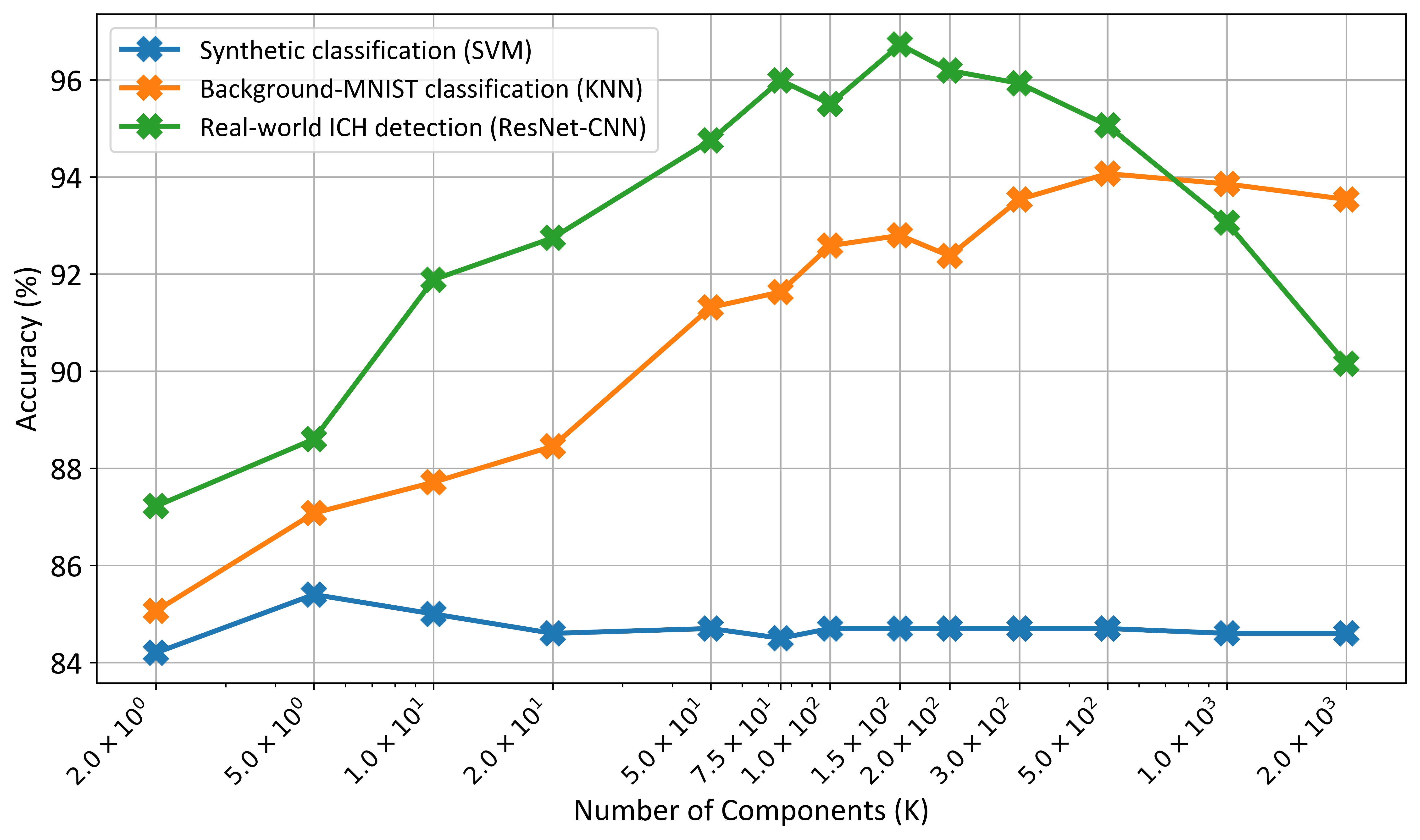}
        \caption{Comparison of the mechanism learning-based deconfounded models' accuracy with a wide range of $K$ under the three classification tasks discussed in the paper. In general, accuracy plateaus once $K$ is sufficiently large indicating a reasonable model complexity.}
        \label{fig: diff K accuracy clas}
        \vspace{-0.5em}
    \end{figure}

    Next, we train three ResNet-CNNs models: (i) a confounded ResNet-CNN, (ii) a CB-based deconfounded ResNet-CNN and (iii) a mechanism learning-based deconfounded ResNet-CNN. Their prediction performances are compared on both the confounded and synthesized non-confounded test sets, as demonstrated in Table~\ref{tab: expr results}. In agreement with theory and synthetic experiments, the mechanism learning-based deconfounded ResNet-CNN, significantly exceeds the performance of the confounded and CB-based deconfounded ResNet-CNNs with the same network architecture by $0.13$ and $0.07$ of F1 score on the synthetic non-confounded test set, respectively. Meanwhile, Figure~\ref{fig: diff K accuracy clas} shows that the optimal $K$ ranges from 75 to 300 in this task.

    Our interpretation of this finding is that the mechanism learning-based ResNet-CNN largely eliminates the effect of unknown confounding factors. By contrast, the confounded model's performance dramatically degrades on the independent synthetic non-confounded test set. Furthermore, mechanism learning demonstrates a clearer, comparable advantage to CB-based method in the real-world, high-dimensional and complex ICH detection task. Note that here, our aim is not to produce the most complex predictor (there could indeed be ML predictors which achieve higher accuracy for this problem); instead the aim of this experiment is to demonstrate the effectiveness of mechanism learning by comparison to classical supervised learning and the CB-based deconfounding method.

\section{Conclusion and Discussion}

In this paper we propose \textit{mechanism learning}, a simple, novel method to train existing supervised ML prediction models which are robust to the spurious associations which are common to real-world observational datasets. The method is based on the presented \textit{causally weighted Gaussian mixture models} (CW-GMMs): Using do-calculus and RKHS-based KDEs applied to the front-door structural causal model, we obtain the front-door weights, which are then used to fit CW-GMMs that encode the front-door interventional distribution. With the CW-GMMs that approximate to that interventional distribution, we then are able to sample from the CW-GMMs, thereby simulating data collected from a controlled experiment. In this way, the ML model makes reverse causal inferences of prediction target from feature data, with only access to the confounded observational data.

Through simulations on semi-synthetic and synthetic data, we demonstrated the robustness of the novel training method to potential confounding, confirming that the method is effective at recovering the direct causal, rather than any spurious, influences. Meanwhile, the presented mechanism learning method has shown better deconfounding ability and sample variability than the existing CB-based deconfounding method in the real-world, high-dimensional and complex ICH dataset. In our view, this represents a significant advance towards the deployment of ML methods that can make both accurate and reliable predictions. In high-stakes applications such as medical decision-making or safety-critical engineering systems, predictive accuracy of ML models is not everything: much more important is robustness/reliability, which can be afforded through the appropriate adaptation of structural causal inference to supervised ML modeling, as we demonstrate here.


\appendix

\section{Data generation processes for semi- and fully-synthetic datasets}
\label{sec: Appendix A}

    \subsection{Synthetic classification data}

    For the synthetic classification data, the cause $Y$ and the (latent) confounder $U$ are (conditional) Bernoulli random variables, whose sample spaces are $\Omega_Y$, $\Omega_U = \{ 1, 2 \}$. The mechanism $Z$ and effect $\boldsymbol{X}$ are (multi-dimensional) Gaussian random variables whose parameters depend on their parent variables. Formally, their dependency is:
        \begin{eqnarray}
            \label{eq: frontdoor clas syn data model}
            U & \sim & \mathit{Bernoulli}\left(p=0.25\right) \nonumber\\
            \left.Y\right|U & \sim & \mathit{Bernoulli}\left(p=q_{y}\left(u\right)\right) \nonumber\\
            \left.Z\right|Y & \sim & \mathit{Normal}\left(\mu=\mu_{z}\left(y\right),\sigma^{2}=1\right) \\
            \left.X_{1}\right|Z,U & \sim & \mathit{Normal}\left(\mu=\mu_{x_{1}}\left(z\right),\sigma^{2}=1\right) \nonumber\\
            \left.X_{2}\right|Z,U & \sim & \mathit{Normal}\left(\mu=\mu_{x_{2}}\left(u\right),\sigma^{2}=1\right). \nonumber
        \end{eqnarray}

    For the confounded training and test sets, the parameters are given below:
        \begin{equation}
            \label{eq: frontdoor clas syn param}       
            \begin{array}{cc}
            q_{y}\left(u\right) = \begin{cases}
                        0.2 & \text{if } u=1\\
                        0.8 & \text{if } u=2
                        \end{cases} &
            \mu_{z}\left(y\right) = \begin{cases}
                        -1.2 & \text{if } y=1\\
                        1.2 & \text{if } y=2
                        \end{cases} \\
            \mu_{x_{1}}\left(z\right) = \begin{cases}
                        -1.8 & \text{if } z<0\\
                        1.8 & \text{if } z>0
                        \end{cases} &
            \mu_{x_{2}}\left(u\right) = \begin{cases}
                        -2.4 & \text{if } u=1\\
                        2.4 & \text{if } u=2
                        \end{cases}.
            \end{array}
        \end{equation}
    
    Five thousand ($5,000$) samples are generated for the confounded training set, whereas $1,000$ samples are generated for the confounded test set. With the confounder $U$, the generated positive samples (class 2) are more likely to concentrate in the positive region of $X_2$, by contrast with the non-confounded data. For the non-confounded test set, the confounding path between $U$ and $Y$ is disconnected by setting $q_{y}\left(u\right) = 0.5$, and $1,000$ samples are generated. Consequently, the samples for two classes are optimally separated by the Bayes' boundary at $X_1 = 0$.

    \subsection{Synthetic regression data}
    \label{subsec: Appendix A.2}

    For the synthetic regression data, all variables are (multi-dimensional) Gaussian whose parameters may depend on their parent variables. Formally, their dependency is:
    \begin{eqnarray}
    \label{eq: frontdoor reg syn data model}
        U &	\sim &	\mathit{Normal}\left(\mu=0,\sigma^{2}=1\right) \nonumber\\
        Y &	= &	a_{0}+a_{1}U+\epsilon_{Y} \nonumber\\
        Z &	= &	bY+\epsilon_{Z} \\
        X &	= &	c_{0}U+c_{1}Z+\epsilon_{X}. \nonumber
    \end{eqnarray}		

    The generated confounded and non-confounded datasets have the same sample sizes as the synthetic classification data. The coefficients $a_0 = 2$, $a_1 = 1$, $b=1.5$, $c_1=1$, $c_0=5$ for the confounded setting and $c_0=0$ for non-confounded setting. $\epsilon_{Y}$, $\epsilon_{Z}$, $\epsilon_{X} \sim Normal\left(\mu=0,\sigma^{2}=1\right)$ are a Gaussian noise term. In this case, the presence of the latent confounder $U$ will lead to a smaller slope for the (linear) regression line. More specifically:
    \begin{eqnarray}
        \hat{y}_\mathrm{conf} & = & 0.15x+1.54+\epsilon_{1} \nonumber\\
        \hat{y}_\mathrm{unconf} & = & 0.67x+\epsilon_{2}.
    \end{eqnarray}

    \subsection{Semi-synthetic background-MNIST data}

    To synthesise the background-MNIST dataset, we modify and embed the original handwriting digit images in the MNIST dataset into a front-door causal setting confounded by the background brightness. The following parametric data generation model is used:
    \begin{eqnarray}
        U & \sim & \mathit{Normal}(0,5) \nonumber \\
        \left.Y\right|U & \sim & \mathit{Bernoulli}\left(p=q_{y}\left(u\right)\right) \nonumber \\
        \left.Z\right|Y & \sim & \mathit{Bernoulli}\left(p=r_{z}\left(y\right)\right)\\
        \left. \boldsymbol{X}\right|Z,U & \sim & \mathit{MNIST}(z,u), \nonumber
    \end{eqnarray}
    where the Bernoulli parameter for $Y$ is a function of the value of the confounder $U$ given by:
        \begin{equation}
            \label{eq: Bernoulli Y param for MNIST}      
            q_{y}(u)=\frac{q^{u}}{q^{u}+\left(1-q\right)^{u}},
        \end{equation}
    where $q = 0.8$ for confounded data. For $q = 0.5$, the data is  non-confounded data because $U$ imposes no effect on $Y$. $\mathit{MNIST}(y,u)$ is a random function that retrieves a unique MNIST image $\boldsymbol{x}$ representing digit ``2'' for $y=1$ and digit ``6'' for $y=2$, and then the brightness of the selected image is converted by $\boldsymbol{x}\mapsto \min \left(\boldsymbol{x} + \boldsymbol{v}(u), 255 \right) $ where $\boldsymbol{\boldsymbol{v}(u)}$ is a same-shape matrix as the MNIST image whose value depends on $u$, that is:
        \begin{equation}
            \label{eq: brightness mapper}  
            v_{ij}(u)=100\times\left(\frac{1}{2}\arctan\left(\frac{1}{5}u\right)+\frac{1}{2}\right),
            \end{equation}
    which maps the continuous confounder values onto the scaled background brightness values in the range $\left[0, 100 \right]$.

\section{Theoretical basis of front-door weights (a recall)}
\label{sec: Appendix B}

For a set of variables that satisfy front-door criterion, we can use \emph{do-calculus} to compute the interventional distribution \citep{pearl2010introduction}:    
        \begin{equation}
        \label{eq: non-marg front-door intv prob}
        \begin{split}
        p\left(\left. x,y',z\right|do\left(y\right)\right) & =  p\left(\left.x\right|y',z\right)p\left(y'\right)p\left(\left.z\right|y\right) \\
        & =  p\left(x,y',z\right)\frac{p\left(\left.z\right|y\right)}{p\left(\left.z\right|y'\right)},
        \end{split}
        \end{equation}
        where the random variable $Y'$ is a copy of variable $Y$ which arises during the application of \emph{do-calculus}.
    
        The joint distribution $p\left(x,y',z\right)$ can be replaced by any \emph{kernel density estimator} (KDE):
        \begin{equation}
            \label{eq: RKHS KDE}
            p\left(x,y',z\right)\thickapprox\frac{1}{N}\sum_{n\in\mathcal{N}}K\left[x-x_{n}\right] K\left[y'-y_{n}\right] K\left[z-z_{n}\right],
        \end{equation}
        where $K\left[\cdot \right]$ is some some kernel density function, $N$ is the sample size, $x_n$, $y_n$ and $z_n$ are realizations of the effect variable $X$, the cause variable $Y$ and the mechanism variable $Z$, respectively. Inserting \eqref{eq: RKHS KDE} into \eqref{eq: non-marg front-door intv prob}, we obtain:
        \begin{align}
        p\left(x, y', z \mid do(y)\right) 
        &\thickapprox \frac{1}{N} \sum_{n \in \mathcal{N}} 
        K\left[x_n - x\right] K\left[z - z_n\right] \nonumber \\
        &\quad \cdot K\left[y' - y_n\right] 
        \frac{p\left(z \mid y\right)}{p\left(z \mid y'\right)}.
        \end{align}
    
        The variables $Y'$ and $Z$ are not of interest to the cause-effect relationship, thus we marginalize them out and interchange the summation and integration by Tonelli’s theorem \citep{folland1999real} due to the non-negativity of the distribution and kernel functions by construction:
        \begin{align}
        p\left(\left.x\right|do\left(y\right)\right) & \thickapprox\frac{1}{N}\sum_{n\in\mathcal{N}}K\left[x_{n}-x\right] \nonumber \\ 
         & \cdot\text{\ensuremath{\underbrace{\int\int K\left[z-z_{n}\right]K\left[y'-y_{n}\right]\frac{p\left(\left.z\right|y\right)}{p\left(\left.z\right|y'\right)}dy'dz}_{\text{front-door causal weight: }w_{n}}}}
        \end{align}
        
        Therefore, the front-door weights are given by:
        \begin{equation}
            \label{eq: weights with integrals}
            w_{n}=\frac{1}{N}\int\int K\left[z-z_{n}\right]K\left[y'-y_{n}\right]\frac{p\left(\left.z\right|y\right)}{p\left(\left.z\right|y'\right)}dy'dz.
        \end{equation}   
        
        The conditional distribution $p\left(\left.z\right|y\right)$ may typically be estimated from the observational data using a suitable estimator, unless prior knowledge is available.
    
        \emph{Reproducing Kernel Hilbert Spaces} (RKHS) functions have the \emph{reproducing property} \cite{aronszajn1950theory}, where:
        \begin{equation}
            \label{eq: RKHS reproducing property}
            \left\langle p,K\left[x,\cdot\right]\right\rangle =\int p\left(x'\right)K\left[x-x'\right]dx'=p(x),
        \end{equation}   
        and, given an RKHS kernel for the KDE, we can use this property to evaluate the inner integrals in \eqref{eq: weights with integrals}. Finally, this means the weights $w_n$ are essentially evaluated at each observational data point:
        \begin{equation}
        \label{eq: front-door cb weights appendix}
        \begin{split}
            w_{n}\left(z_n, y_n, y\right) & =  \left.\frac{p\left(\left.z\right|y\right)}{Np\left(\left.z\right|y'\right)}\right|_{z=z_{n},y'=y_{n}} \\
             & =  \frac{p\left(\left.z_{n}\right|y\right)}{N p\left(\left.z_{n}\right|y_{n}\right)}.
        \end{split}
        \end{equation}

\end{document}